\documentclass{article}

\PassOptionsToPackage{sort, numbers}{natbib}

\usepackage[final]{neurips_2023}

\usepackage[utf8]{inputenc} 
\usepackage[T1]{fontenc}    
\usepackage{hyperref}       
\usepackage{url}            
\usepackage{booktabs}       
\usepackage{amsfonts}       
\usepackage{amsmath}
\usepackage{amssymb}
\usepackage{nicefrac}       
\usepackage{microtype}      
\usepackage{xcolor}         
\usepackage{xspace}         
\usepackage{graphicx}
\usepackage{subcaption}
\usepackage{bm}
\usepackage{physics}
\usepackage[capitalize]{cleveref}       
\usepackage{multirow}
\usepackage[font=small]{caption}
\usepackage[frozencache,cachedir=.]{minted}

\newcommand{\sdrop}{\textsc{SparseDrop}\xspace}

\newcommand{\m}[1]{\mathbf{\bm{#1}}}
\newcommand{\R}{\mathbb{R}}
\newcommand{\T}{\top}

\DeclareMathOperator{\bern}{Bernoulli}
\DeclareMathOperator{\gemm}{GEMM}

\hypersetup{
    colorlinks,
    linkcolor={red!50!black},
    citecolor={blue!50!black},
    urlcolor={black}
}

\title{Efficient Sparse Training with Structured Dropout}

\author{%
  Andy Lo \\
  Computer Laboratory\\
  University of Cambridge\\
  \texttt{cyal4@cam.ac.uk} \\
}

\begin{document}

\maketitle

\begin{abstract}
Dropout is a common regularisation technique in deep learning that improves generalisation. Even though it introduces sparsity and thus potential for higher throughput, it usually cannot bring speed-ups on GPUs due to its unstructured nature. In this project, I experiment with \sdrop, a structured, hardware-friendly variant of dropout that can exploit such sparsity. I provide a CUDA implementation of \sdrop, achieving speed-ups against its dense counterpart even at low sparsity levels. The empirical results demonstrate that \sdrop provides similar, or sometimes even better, regularisation properties as standard dropout. This suggests its potential as a drop-in replacement to standard dropout with faster training speeds. The source code is available at \url{https://github.com/andylolu2/sparse-dropout}.
\end{abstract}

\section{Introduction}

Designing machine learning algorithms that can be executed efficiently on modern hardware is core to the success of many prior efforts \citep{hardware-lottery, lecun-cnn, alexnet}. A popular area of research is to utilise sparsity to improve training and/or inference efficiency by skipping over unnecessary computation \citep{han2015learning,sparse-finetuning,optimal-brain-damage,scnn}. While sparse computation has successfully brought substantial speed-ups on CPUs \citep{sparse-finetuning}, its typically unstructured nature makes it difficult to execute efficiently on highly parallelised hardware like GPUs. To tackle this issue, structured pruning is an emerging method for accelerating inference on GPUs. Unlike most prior works, this paper focuses on the use of structured sparsity for \emph{GPU training}.

Unlike most prior works that motivate dropout by performance, this project studies dropout from a sparse computation perspective. I ask the question: \emph{Is it possible to exploit the sparsity of dropout for faster computation while maintaining the same regularisation properties?} Experience from prior works suggests that we must introduce sparsity with structure to achieve practical computation speed-ups. In light of this, this paper proposes \sdrop, a simple, structured, and hardware-friendly variant of dropout that can benefit from sparsity on GPUs. Concretely, the previous and/or next operation of the dropout can now become a (structured) sparse operator, effectively fusing the dropout and the previous/next layer. In this paper, I will only focus on optimising the specific pattern of `dropout followed by matrix multiplication', leaving other patterns as future work.

Achieving the theoretical benefits of \sdrop is a challenging task as it requires low-level control over GPU computation and memory. Such fine-grained control over the hardware is not exposed by common deep learning libraries such as TensorFlow \citep{tensorflow}, PyTorch \citep{pytorch}, or JAX \citep{jax}. Instead, I implemented several sparse matrix multiplication kernels in CUDA whose execution time decreases linearly with the sparsity level. \Cref{sec:method} describes the important implementation details to achieve similar or better throughput compared to dense matrix multiplication. The experimental results in \cref{sec:exp} reveal that \sdrop matches the performance of standard dropout. On some tasks, it even converges to \emph{better solutions with less training time}.

\section{Background}

This section aims to provide sufficient background knowledge for understanding the design choices made in this project. \cref{background:dropout,background:gemm} introduce the basics of dropout and efficient matrix multiply. \cref{background:related} discusses how this paper relates to several prior works. 

\subsection{Dropout} \label{background:dropout}

Dropout is a stochastic regularisation technique that helps to learn more robust features and has been shown to improve generalisation across many domains \citep{dropout, transformer}. The standard dropout applied to an activation matrix $\m{X} \in \R^{M \times K}$ randomly drops each value with hyper-parameter probability $p$. Since dropout is only used during training and becomes a no-op during inference, additional scaling is applied to un-dropped values to avoid train-inference mismatch. The forward and backward pass are defined as:
\begin{equation*}
    \begin{split}
        \m{m} \sim \bern(1-p)^{M \times K}
        \qquad
        \bar{\m{X}} = \frac{1}{1-p} \m{X} \odot \m{m}
        \qquad
        \pdv{L}{\m{X}} = \frac{1}{1-p} \pdv{L}{\bar{\m{X}}} \odot \m{m}
    \end{split}
\end{equation*}
where $\odot$ denotes element-wise multiplication.

The reasons why dropout is a good regularisation method are beyond the scope of this project, hence the theoretical background is omitted here. This project will instead take an empirical approach and test the effectiveness of dropout by experiments.

\subsection{Matrix multiplication on GPU} \label{background:gemm}

Matrix multiplication (also known as GEMM\footnote{GEMM is the acronym for \underline{GE}neral \underline{M}atrix-to-matrix \underline{M}ultiply}) is one of the most foundational operations in deep learning, commonly referred to as a `Dense' layer in neural networks. Consequently, many years of hardware and software engineering efforts have been put into making GEMM as efficient as possible. Here, we cover the fundamentals of implementing efficient dense GEMM which will help later as we implement performant sparse GEMM.

Efficient GEMM is designed to maximise the use of hardware features. GPUs thrive at parallel computation hence it is necessary to partition the GEMM problem into smaller, independent sub-problems that can be solved concurrently. The highest level of parallelism is across \emph{threadblocks}, a group of threads that executes on the same \emph{streaming multiprocessor}. Consider the problem $\m{C} = \m{A} \m{B}$ where $\m{A} \in \R^{M \times K}$, $\m{B} \in \R^{K \times N}$, and $\m{C} \in \R^{M \times N}$. We say this is a problem of size $(M, N, K)$. Consider a partition of the problem in the $M$ and $N$ dimension by block sizes $M_{blk}$ and $N_{blk}$ respectively\footnote{We assume the block sizes divide the problem sizes in this paper to simplify the implementation.}:
\begin{equation*}
    \m{A} = \begin{bmatrix}
            \m{A}_0 \\
            \vdots \\
            \m{A}_{M / M_{blk}}
             \end{bmatrix}
    \quad
    \m{B} = \begin{bmatrix}\m{B}_0 & \dots & \m{B}_{N / N_{blk}}\end{bmatrix}
    \quad
    \m{C} = \begin{bmatrix}
                \m{A}_0 \m{B}_0 & \dots & \m{A}_0 \m{B}_{N / N_{blk}} \\
                \vdots      & \ddots & \vdots \\
                \m{A}_{M / M_{blk}} \m{B}_0 & \dots & \m{A}_{M / M_{blk}} \m{B}_{N / N_{blk}} \\
             \end{bmatrix}
\end{equation*}
where $\m{A}_i \in \R^{M_{blk} \times K}$ and $\m{B}_j \in \R^{K \times N_{blk}}$. This gives $(M / M_{blk}) \cdot (N / N_{blk})$ independent sub-problems of size $(M_{blk}, N_{blk}, K)$, each of which is computed by one threadblock. For a large $K$, it is not possible to load $\m{A}_i$ or $\m{B}_j$ into lower-level memory entirely. A typical implementation will thus also split the $K$ dimension into blocks of size $K_{blk}$ and accumulate over the partial sums:
\begin{equation*}
    \begin{split}
        \m{A}_i = \begin{bmatrix}\m{A}_{i,0}, \dots, \m{A}_{i, K / K_{blk}}\end{bmatrix}
        \qquad
        \m{B}_j = \begin{bmatrix}
                \m{B}_{0, j} \\
                \vdots \\
                \m{B}_{K / K_{blk}, j}
                 \end{bmatrix}
        \qquad
        \m{A}_i \m{B}_j = \sum_{k=0}^{K / K_{blk}} \m{A}_{i, k} \m{B}_{k, j}
    \end{split}
\end{equation*}
where $\m{A}_{i,k} \in \R^{M_{blk} \times K_{blk}}$ and $\m{B}_{k,j} \in \R^{K_{blk} \times N_{blk}}$. 

To implement an efficient kernel, we must first understand the typical bottlenecks of GPU computation. One important property of is that global memory (VRAM) is comparatively slow, meaning that most operations, including GEMM, are \emph{global memory bound}. To alleviate the problem, $\m{A}_{i,k}$ and $\m{B}_{k,j}$ are first loaded into the \emph{shared memory} for each threadblock. Subsequent reads of the data are then served by the shared memory directly, improving data reuse and thus reducing the load on the global memory. Nonetheless, global memory remains the main bottleneck to faster GPU algorithms. More performant algorithms are only possible if they reduce global memory accesses \citep{flash-attention, mamba}.

I would like to emphasise that efficient GEMM is a complicated algorithm and even achieving on-par performance with standard libraries such as PyTorch (which uses CuBLAS) is a challenging task. To reduce the complexity of the implementation, libraries like CUTLASS \citep{cutlass} exist to make high-performance GEMM more customisable. The high ceiling of CUTLASS comes at the cost of a steep learning curve which adds to the engineering difficulty of this project. 

\subsection{Related work} \label{background:related}

Structured dropout have been explored in prior works \citep{drop-block,drop-cluster,batch-drop-block,revisiting}. However, all of them are motivated by performance (e.g., accuracy). For example, DropBlock \citep{drop-block} identifies that standard dropout performs poorly when used with convolutional neural networks \citep{lecun-cnn} and demonstrates that better generalisation can be achieved if consecutive pixels (patches) of an image are dropped together. This makes it an exciting line of research since structured dropout can be favourable from both a performance and, as shown later, computational perspective.

There is also a long lineage of work on training sparse neural networks \citep{optimal-brain-damage}. However, most work focuses on faster or lower-memory inference rather than speeding up training. In fact, it is common to use \emph{more} training resources for sparse networks. For example, all of \citep{accelerating-snn,training-snn,sparse-finetuning,han2015learning,scnn,scnn-slimming} first fully train a dense network and then apply pruning with re-training to obtain a performant sparse model. \citep{conditional-computation} is the most similar to the idea presented in this project, suggesting to use block-sparse patterns to accelerate GPU training. However, the implementation\footnote{\url{https://github.com/bengioe/condnet/}} is rather out-of-date and the authors were only able to achieve speedups at high sparsity levels ($\geq 75\%$).

\section{Method} \label{sec:method}

\subsection{Fusing dropout with GEMM}

\begin{figure}[t]
    \centering
    \subcaptionbox
    {Unstructured \texttt{dsd\_matmul}. Since data is accessed in blocks, all blocks still needs to be loaded. \label{fig:dropout}}
    [.32\textwidth]
    {\includegraphics[width=\linewidth]{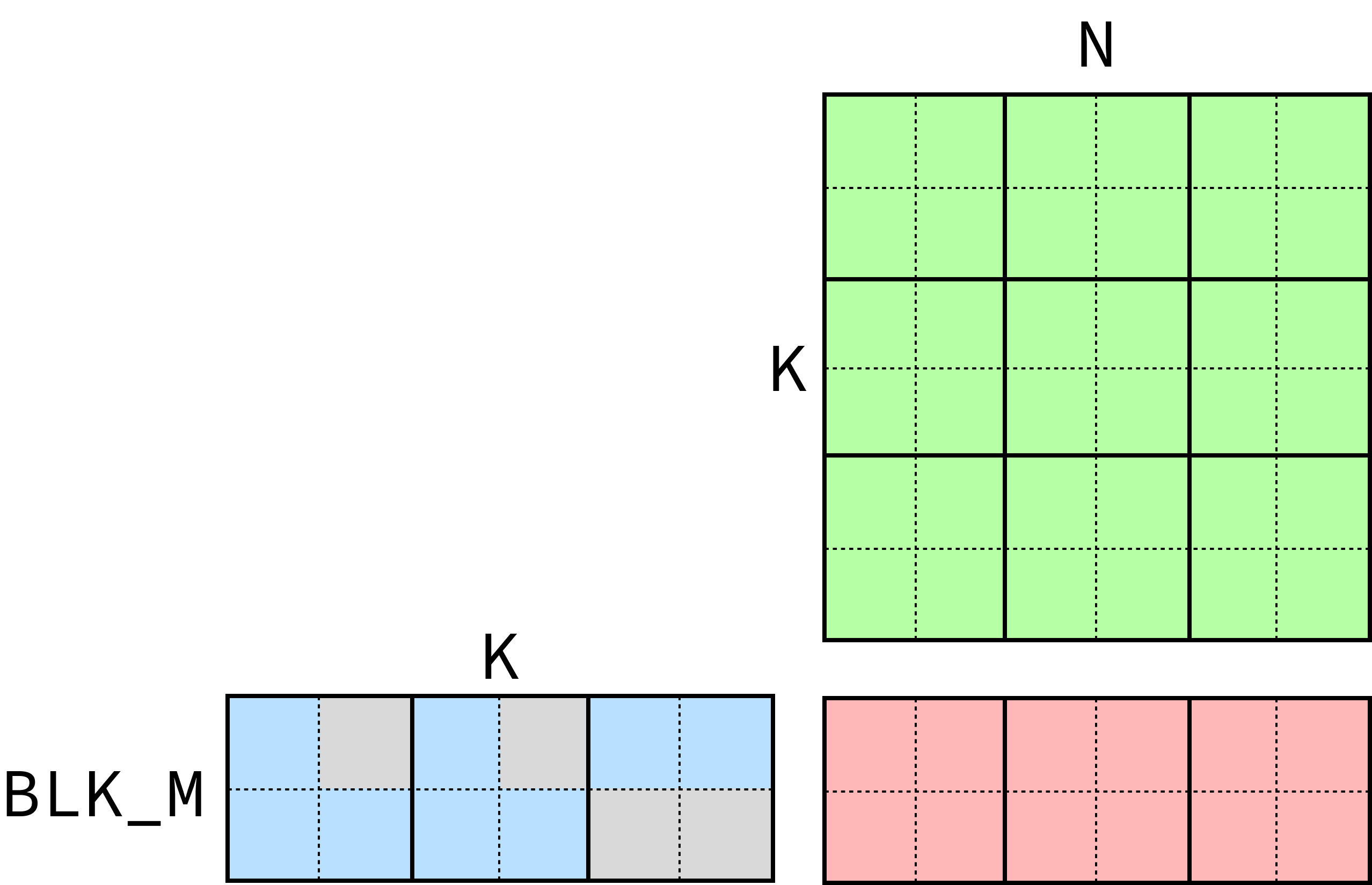}}%
    \hfill
    \subcaptionbox
    {Structured \texttt{dsd\_matmul}. The second block of the first input (blue) are zeros, thus the entire second row-block of the second input (green) can be ignored. \label{fig:sdropout-dsd}}
    [.32\textwidth]
    {\includegraphics[width=\linewidth]{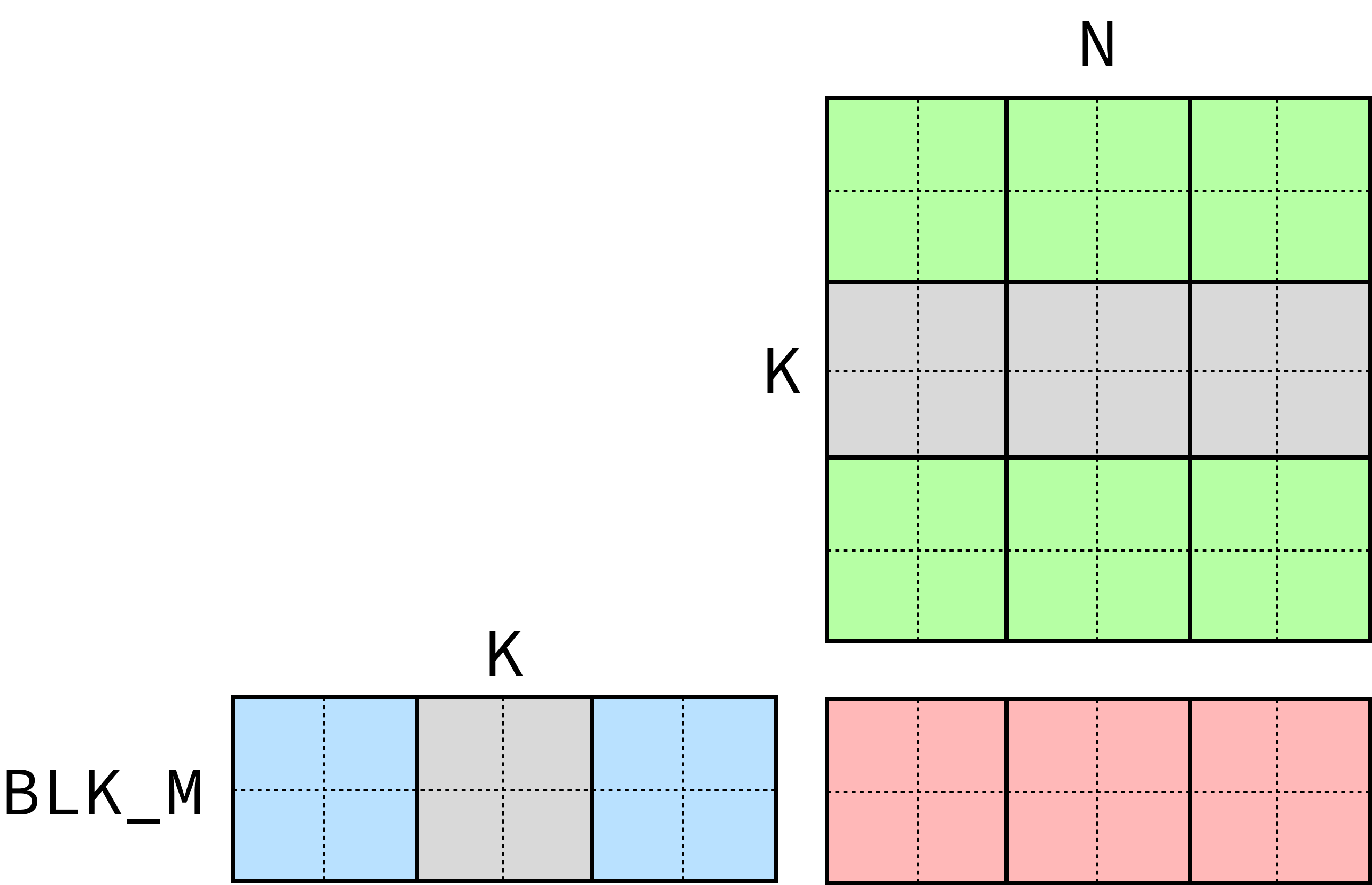}}%
    \hfill
    \subcaptionbox
    {Structured \texttt{sdd\_matmul}. The second block of the output (red) are zeros, thus the entire second column-block of the second input (green) can be ignored. \label{fig:sdropout-sdd}}
    [.32\textwidth]
    {\includegraphics[width=\linewidth]{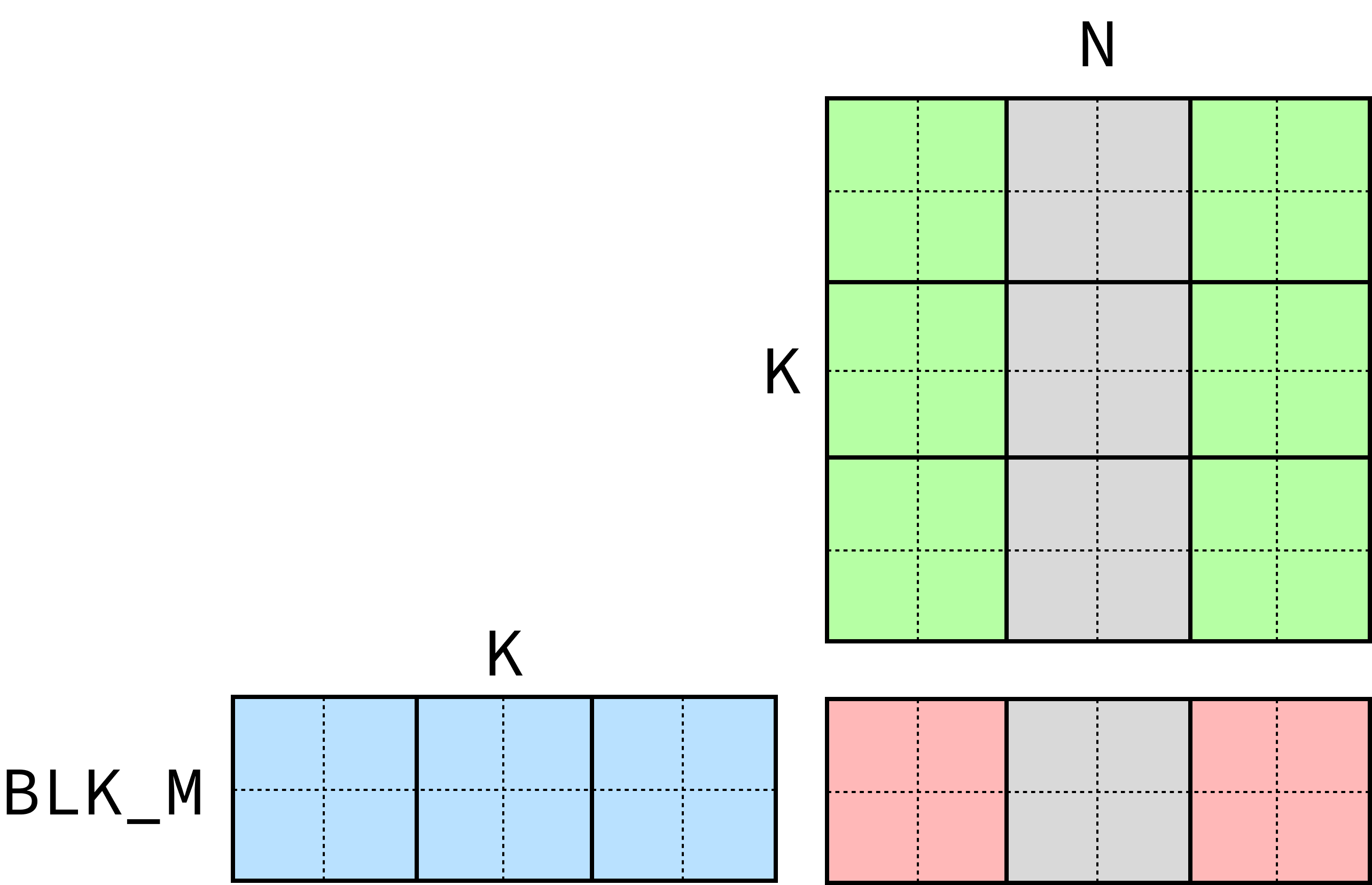}}%
    \caption{Standard versus block-wise sparse GEMM for one $M$-block with $M_{blk} = N_{blk} = K_{blk} = 2$. The blue matrix represents a fragment of input $\m{A}$, the green matrix is input $\m{B}$ and the red matrix is the output $\m{C}$. Solid lines represent the block-wise access granularity by the GPU. Ignored elements are coloured in gray. The sparsity level is 33\% in all cases but only the structured variants can benefit from it by skipping over entire blocks.}
    \label{fig:dropout-comparison}
\end{figure}

To benefit from sparsity introduced by dropout, we must replace dense operators before/after the dropout layer with their sparse counterparts. As discussed earlier, we will limit our scope to optimising the `dropout + linear layer` pattern. To further simplify the implementation, we will only consider linear layers without the bias term.\footnote{There is some evidence suggesting that the bias term is not always necessary for good performance. \citep{nanogpt}} The forward and backward formulae are:
\begin{align}
    \m{m} & \sim \bern(1-p)^{M \times K} \nonumber \\
    \m{Y} &= \frac{1}{1-p} (\m{X} \odot \m{m}) \m{W} \label{eq:sparse_y} \\
    \pdv{L}{\m{X}} &= \frac{1}{1-p} \left( \pdv{L}{\m{Y}} \m{W}^\T \right) \odot \m{m} \label{eq:sparse_dx} \\
    \pdv{L}{\m{W}} &= \frac{1}{1-p} (\m{X} \odot \m{m})^\T \pdv{L}{\m{Y}} \label{eq:sparse_dw}
\end{align}
where $\m{X} \in \R^{M \times K}$ and $\m{W} \in \R^{K \times N}$. We see that \cref{eq:sparse_y} and \cref{eq:sparse_dw} is a GEMM (with scaling) that produces a \underline{d}ense output from one \underline{s}parse input and one \underline{d}ense input, commonly abbreviated as a \texttt{dsd\_matmul}. On the other hand, \cref{eq:sparse_dx} produces a \underline{s}parse output from two \underline{d}ense inputs, abbreviated as an \texttt{sdd\_matmul}. As shown in \cref{fig:dropout}, the unstructured sparsity pattern defined by $\m{m}$ makes it difficult to accelerate in hardware in the general case.

\subsection{\sdrop}

To make the sparse GEMM fast in hardware, I propose to \emph{alter the semantics to match the hardware requirements}. In particular, \sdrop uses a block-sparse masking matrix $\m{m'}$ in place of the per-element $\m{m}$ while keeping \cref{eq:sparse_y,eq:sparse_dx,eq:sparse_dw} the same. The important implementation detail is to \emph{choose the same block sizes (i.e. $M_{blk}$ and $K_{blk}$) for $\m{m'}$ and the matrix multiplication algorithm}. I found that it is often a better strategy to prioritise $M_{blk}$ and $K_{blk}$ for GEMM efficiency over the masking granularity as choosing the wrong block sizes incurs a significant performance penalty. 

\cref{fig:sdropout-dsd} depicts how the block-sparse mask allows us to accelerate the \texttt{dsd\_matmul} kernel. For each threadblock, instead of iterating through all $K / K_{blk}$ blocks from the inputs, we can now skip over entire blocks that have been logically masked by $\m{m'}$. Crucially, such implementation allows us to \emph{reduce the global memory bottleneck} since blocks that are not required do not need to be read from global memory at all, meeting the critical requirement for faster GPU kernels as discussed in \cref{background:gemm}.

Similarly, the \texttt{sdd\_matmul} kernel can be accelerated by only computing the output blocks that have not been masked (assuming the output matrix has been initialised to zeros). A simple example is shown in \cref{fig:sdropout-sdd}.

\subsection{Block splitting}

\begin{figure}[t]
    \centering
    \subcaptionbox
    {Logical mask}
    [.27\textwidth]
    {\includegraphics[width=\linewidth]{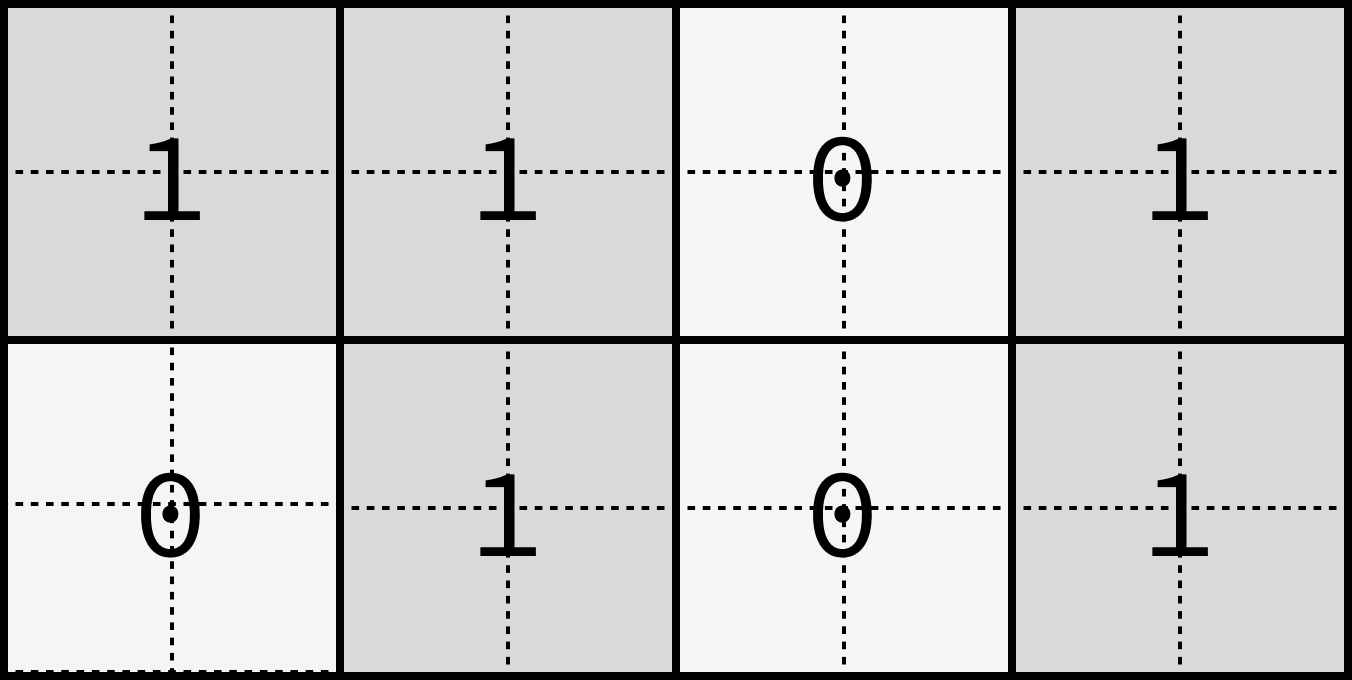}}%
    \hfill
    \subcaptionbox
    {Retiled mask ($p=2,q=1$)}
    [.27\textwidth]
    {\includegraphics[width=\linewidth]{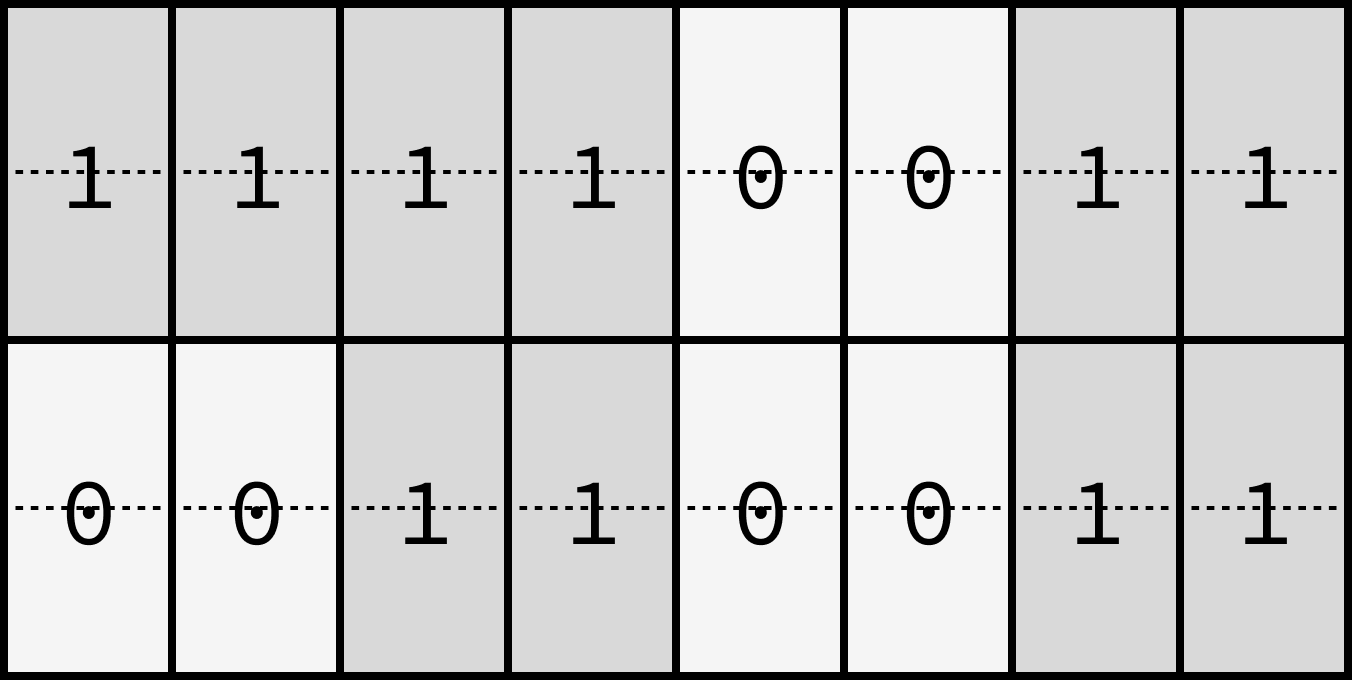}}%
    \hfill
    \subcaptionbox
    {Retiled mask ($p=1,q=2$)}
    [.27\textwidth]
    {\includegraphics[width=\linewidth]{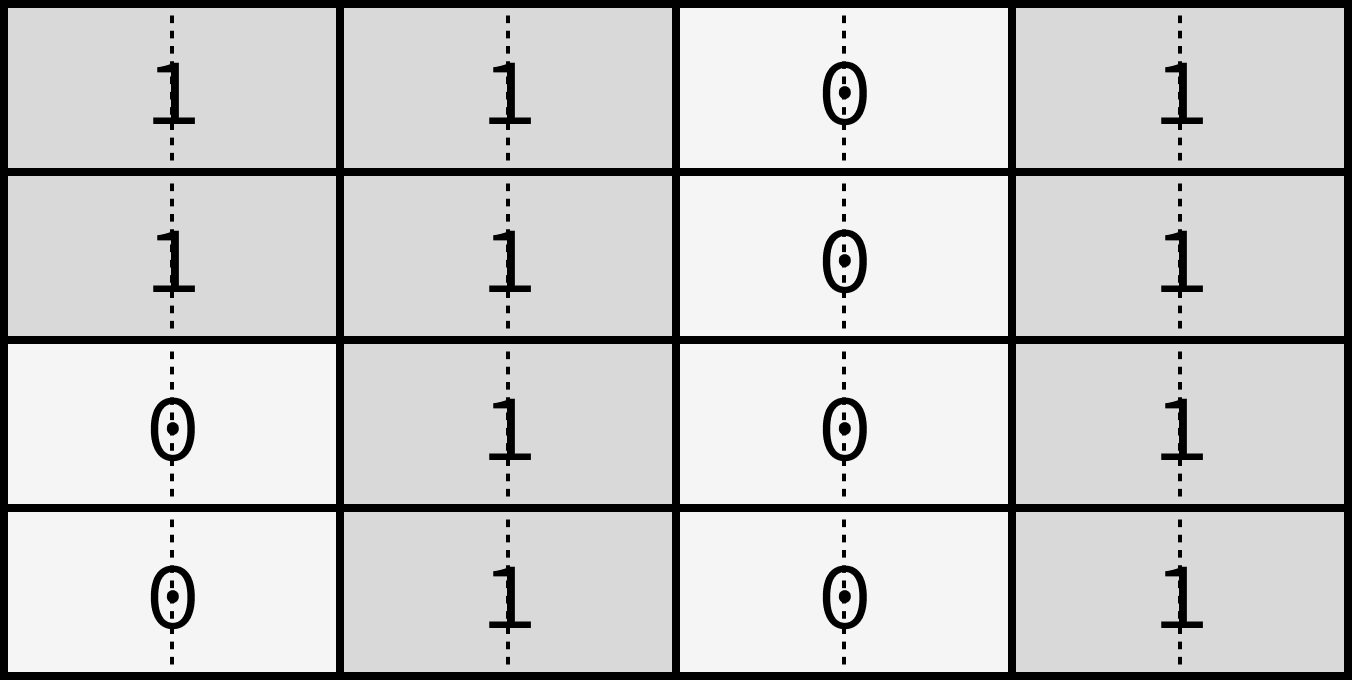}}%
    \caption{Example of block splitting. The logical block size is $2 \times 2$ but mask (b) operates on $2 \times 1$ while (c) operates on $1 \times 2$. The semantics is the same across all three masks.}
    \label{fig:mask-splitting}
\end{figure}

One limitation of choosing the same block sizes for $\m{m'}$ and GEMM is that it ties the choice of block sizes for the forward and backward pass. We denote the GEMM algorithm for a $(M, N, K)$ problem with block sizes of $(M_{blk}, N_{blk}, K_{blk})$ as $\gemm(M,N,K,M_{blk},N_{blk},K_{blk})$. If the forward pass \cref{eq:sparse_y} is implemented as $\gemm(M,N,K,M_{blk},N_{blk},K_{blk})$, \cref{eq:sparse_dw} then necessarily needs to be a $\gemm(K,N,M,K_{blk},N_{blk}',M_{blk})$. This is problematic since the optimal choices of $M_{blk}$, $K_{blk}$ for the forward pass might be sub-optimal for the backward pass. I observe that this can sometimes lead to 2-10$\times$ worse throughput.

To alleviate this problem, I propose to perform \emph{block splitting}. Given a logical block-wise mask with block sizes $(M_{blk}, K_{blk})$, we can retile the mask with a smaller block size $(M_{blk} / p, K_{blk} / q)$ and repeat each entry $p$ times horizontally and $q$ times vertically to obtain a logically equivalent mask, assuming $p$ and $q$ divides $M_{blk}$ and $K_{blk}$ respectively. An illustration is shown in \cref{fig:mask-splitting}. This now allows us to implement \cref{eq:sparse_y} as $\gemm(M,N,K,M_{blk},N_{blk},K_{blk}/2)$ (by choosing $p=1,q=2$) and \cref{eq:sparse_dw} as $\gemm(K,N,M,K_{blk},N_{blk}',M_{blk}/2)$ (by choosing $p=2,q=1$) which has proven to provide sufficient flexibility to recover most of the performance. 

\subsection{Implementation details}

To ensure optimal GEMM performance, most of the typical dense GEMM optimisations need to be implemented. These include the use of tensor cores \citep{tensor-cores}, vectorised memory accesses \citep{vectorized-memory}, threadblock swizzling \citep{threadblock-swizzle} and shared memory layout swizzling \citep{cuda-docs}. I empirically observed that each of such optimisations provided at least 20\% improvement in throughput. To simplify the implementation, I specialise the kernel for NVIDIA Turing architecture (specifically RTX 2060 Max-Q), ignoring any cross-platform or cross-generation compatibility concerns. Even with the help of CUTLASS, the \texttt{dsd\_matmul} and \texttt{sdd\_matmul} takes $\approx 650$ lines of CUDA code to implement.

I also found that a naive PyTorch implementation of sampling $\m{m}'$ incurs significant overhead. This is because \cref{eq:sparse_y,eq:sparse_dx,eq:sparse_dw} each requires $\m{m}'$ in a different format and doing the conversion is non-trivial. Using the PyTorch profiler, I examined the stack trace for problems of various sizes and discovered that for small-to-medium sized problems (e.g., $M,N,K \leq 1024$), most of the time is spent on generating the mask $\m{m}'$ rather than GEMM. To close the performance gap, I re-implemented the mask generation in C++ with a Python binding. This custom implementation also allowed me to efficiently pack the mask bits as 64-bit integers which further reduced the overhead of the sparse kernel\footnote{PyTorch currently does not have a bit-packing feature, so a naive implementation would require one global memory read per inner iteration.}. 

\subsection{Performance benchmarking} \label{sec:benchmark}

To test the performance of \sdrop, we compare it against several baseline implementations. They are dense GEMM (\textbf{Dense}), standard dropout followed by dense GEMM (\textbf{Dropout + Dense}), and block-wise dropout followed by dense GEMM (\textbf{Block dropout + Dense}) implemented in PyTorch\footnote{At the time of writing, PyTorch has an experimental feature for block-sparse GEMM. However, the backward pass for block-sparse GEMM has not been implemented yet thus it has not been considered as a baseline.}.

\begin{figure}[t]
    \centering
    \subcaptionbox
    {Total time (forward + backward) for $M=N=K=1024$ at various sparsity levels. \label{fig:benchmark-time}}
    [.48\textwidth]
    {\includegraphics[width=\linewidth]{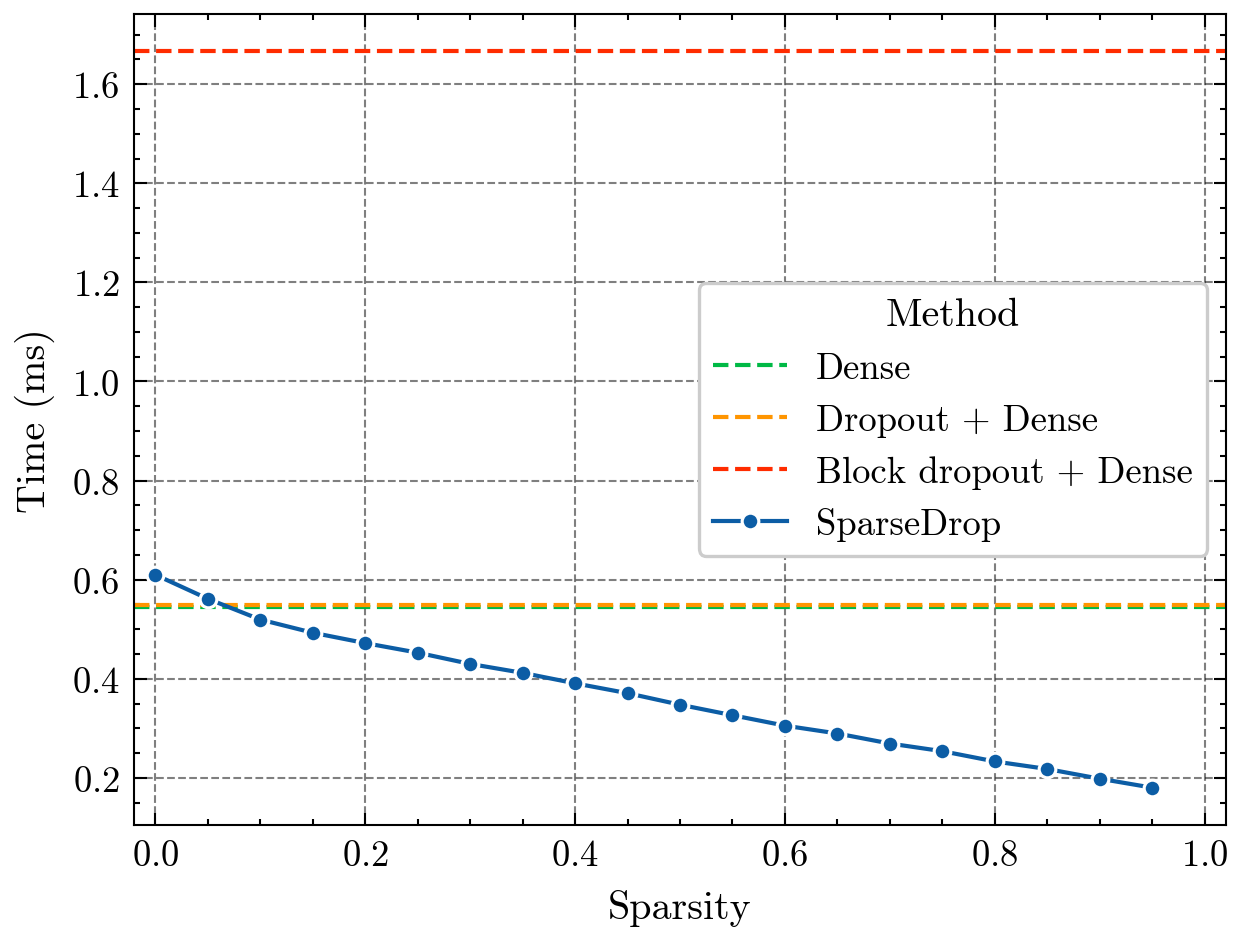}}%
    \hfill
    \subcaptionbox
    {FLOPS achieved by \sdrop for $M=N=K=1024$ at various sparsity levels. The dotted line represents the minimum FLOPS required to achieve speed-up over \textbf{Dense}. \label{fig:benchmark-flops}}
    [.48\textwidth]
    {\includegraphics[width=\linewidth]{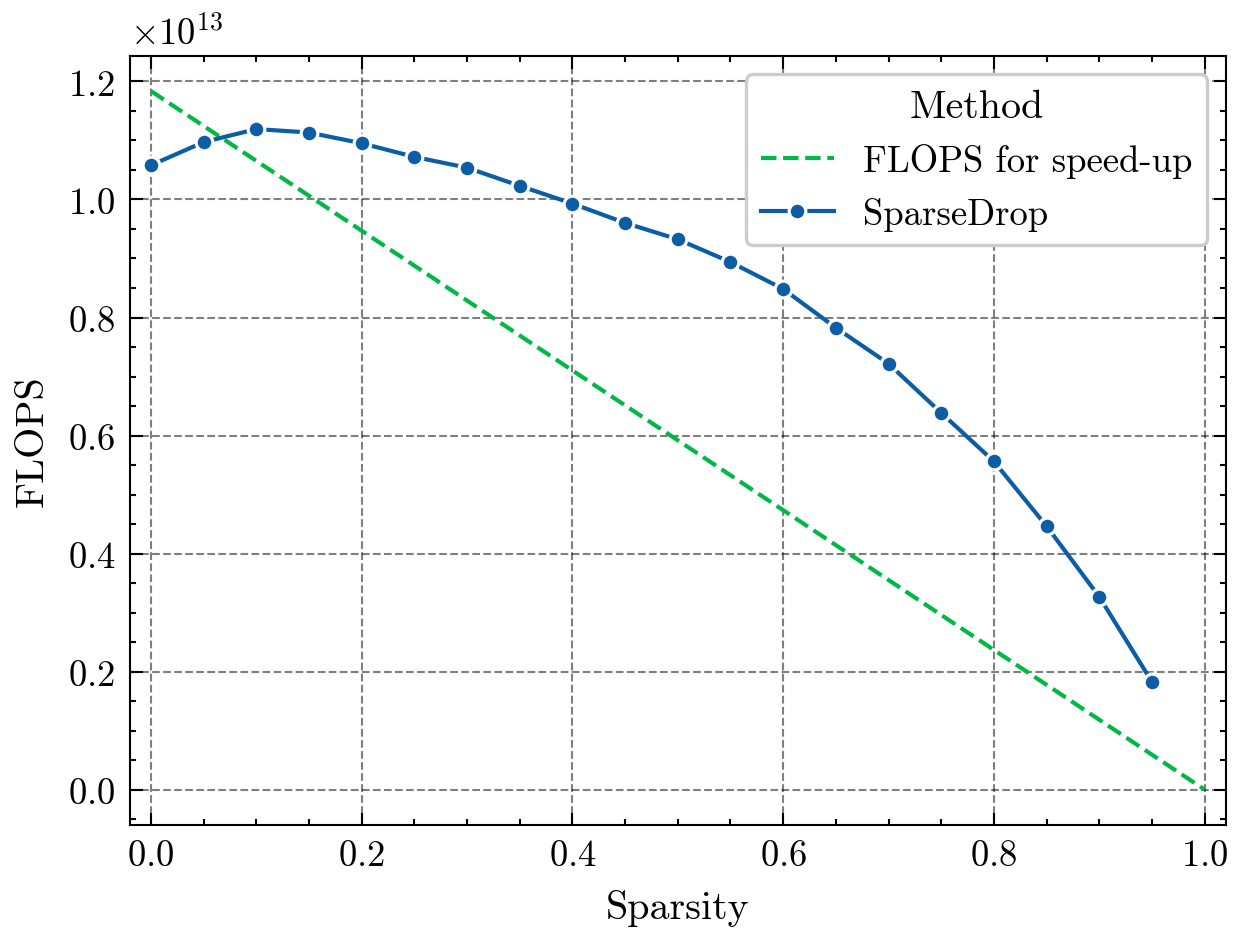}}%
    \caption{Benchmark of \sdrop against baseline methods, measured on RTX 2060 Max-Q with GPU clock locked at 1200MHz and clearing the L2 cache between each measurement.}
    \label{fig:benchmark}
\end{figure}

\cref{fig:benchmark} demonstrates that \sdrop can achieve speed-up over the baseline methods even at low sparsity levels. \cref{fig:benchmark-time} shows that \textbf{Block dropout + Dense} is significantly slower than the rest of the methods, highlighting the bottleneck from naively generating and applying the mask using PyTorch. At sparsity levels $> 5\%$, \sdrop already executes faster than \textbf{Dense} and \textbf{Dropout + Dense} and the execution time decreases linearly with the sparsity level. This is a significant improvement against other sparsity schemes as many only aim to achieve speed-ups with high sparsity \citep{blocksparse-cuda}. \cref{fig:benchmark-flops} show that \sdrop can maintain a relatively high throughput across many sparsity levels. Counter-intuitively, the FLOPS \emph{increases} slightly with small amounts of sparsity ($\leq 30\%$) as seen in \cref{fig:benchmark-flops}. I fail to find a simple explanation for this phenomenon and leave further investigation as future work. 

\section{Experiments} \label{sec:exp}

This section aims to test the empirical effectiveness of \sdrop. We want to understand whether \sdrop can aid generalisation similar to standard dropout and evaluate its practicality in real-world training. To do so, I apply \sdrop to Multi-Layer Perceptron (MLP) and Transformer \citep{transformer} and evaluate its generalisation performance across both computer vision and natural language tasks in \cref{sec:tasks}. Across all tasks, the block size of \sdrop is fixed to $M_{blk}=128$, $K_{blk}=128$ as it gives good throughput at input size $\approx 1024$. 

The empirical results shown in \cref{sec:result} suggest that \sdrop is competitive against standard dropout both semantically and computationally. In some tasks, it even leads to better generalisation while also training faster. 

\subsection{Tasks and methodology} \label{sec:tasks}

The following sections give a brief overview of each task and the experiment setup. The full description for each experiment can be found in \cref{sec:exp-details}.

\subsubsection{MLP} \label{sec:mlp}

An MLP model is applied to solve image classification tasks. To test the effects of dropout, each linear layer in the standard MLP is replaced with either \textbf{Dense} (equivalent to removing the bias term), \textbf{Dropout + Dense} or \sdrop. The model has one input layer, two hidden layers, one output layer, uses ReLU activations, and has hidden dimension 1024.

As dropout is typically the most effective when a model is prone to overfitting, we only focus on a simple dataset, namely MNIST \citep{mnist}. To exaggerate the effects of overfitting, only 16,384 images are used from the training set. The raw images are resized to $32 \times 32$ to simplify the implementation.

I perform a hyper-parameter search for $p \in \{0, 0.1, 0.2, 0.3, 0.4, 0.5, 0.6, 0.7\}$ to find the dropout rate for the best generalisation. Each model is trained until the validation accuracy does not increase for 5 consecutive validation checkpoints or up to 100 epochs. \cref{table:mlp-mnist} shows the best $p$ found for each method according to the validation accuracy, averaged across 3 runs for each $p$.

\subsubsection{Vision Transformer}

This section tests \sdrop with a more capable, closer to state-of-the-art architecture: Vision Transformers (ViT) \citep{vit}. On a high level, ViT group patches of pixels into `vision tokens' and apply a standard transformer model to process to token sequence. Similar to \cref{sec:mlp}, each linear layer in ViT is substituted with either \textbf{Dense}, \textbf{Dropout + Dense} or \sdrop. The model uses patch size $2 \times 2$ and has two layers with hidden dimension 1024.

The models are tested on Fashion MNIST \citep{fashion-mnist} and CIFAR-10 \citep{cifar10} with the same pre-processing steps as \cref{sec:mlp}. The other experimental details are the same\footnote{For CIFAR-10, the early-stopping patience is increased to 10 consecutive validation steps as I observe higher variability in the validation accuracy/loss.} as \cref{sec:mlp} except that, due to limited compute, the experiments are not repeated.

\subsubsection{Language modelling}

This section aims to test \sdrop on a different domain: Natural language processing. I focus on language modelling using a GPT-style, decoder-only transformer model. The experiments use standard architecture choices following \citep{gpt2} and \citep{nanogpt} with the exception of the linear layer substitutions. The model consists of 4 layers with hidden dimension 1024.

A character-level model is used to learn the Shakespeare dataset \citep{shakespeare} with 1,115,394 characters. To reduce experiment time, only the first 524,288 tokens are used for training. Since this is not a classification task, the early stopping is performed according to the validation loss.

\subsection{Results} \label{sec:result}

\begin{table}[t]
\caption{Best dropout rate for each method. The accuracy and loss are calculated from the best checkpoint (according to the quantity monitored by early stopping) throughout training. Note that the training time is the end-to-end time until early stopping, thus measuring both the throughput and how quickly the models converge.}
\label{table:mlp-mnist}
\centerline{
\begin{tabular}{ l l l l l l l } 
    \toprule
    Model & Dataset & Method & Best $p$ & Val accuracy
    & Val loss & Training time (minutes)
    \\
     
    \midrule
     
    \multirow{3}{*}{MLP} & \multirow{3}{*}{MNIST} &
    Dense & - &
    96.96{\footnotesize$\pm$0.62} & 0.166{\footnotesize$\pm$0.087} &
    \textbf{1.94{\footnotesize$\pm$0.10}}
    \\
    && Dropout + Dense & 0.5 &
    \textbf{97.82{\footnotesize$\pm$0.24}} & \textbf{0.080{\footnotesize$\pm$0.023}} &
    2.78{\footnotesize$\pm$0.51}
    \\
    && \sdrop & 0.3 &
    97.61{\footnotesize$\pm$0.20} & 0.092{\footnotesize$\pm$0.036} &
    3.18{\footnotesize$\pm$0.13}
    \\
    
    \midrule
    
    \multirow{6}{*}{ViT} & \multirow{3}{*}{Fashion MNIST} &
    Dense & - & 85.69 & 0.419 & \textbf{20.15}
    \\
    && Dropout + Dense & 0.5 & 86.84 & 0.396 & 38.08
    \\
    && \sdrop & 0.2 & \textbf{87.04} & \textbf{0.382} & 22.62
    \\
    
    \cmidrule{2-7}
    
    & \multirow{3}{*}{CIFAR-10} &
    Dense & - & 48.71 & 1.669 & \textbf{12.3}
    \\
    && Dropout + Dense & 0.4 & 53.05 & 1.524 & 24.73
    \\
    && \sdrop & 0.4 & \textbf{56.79} & \textbf{1.264} & 45.05
    \\
    
    \midrule
    
    \multirow{3}{*}{GPT} & \multirow{3}{*}{Shakespeare} &
    Dense & - & - & 1.643 & \textbf{6.48}
    \\
    && Dropout + Dense & 0.6 & - & 1.448 & 46.8
    \\
    && \sdrop & 0.5 & - & \textbf{1.430} & 42.6
    \\
    
    \bottomrule
\end{tabular}
}
\end{table}

The experiment results shown in \cref{table:mlp-mnist} clearly demonstrate that \sdrop has the same regularisation effects as standard dropout. Across all four tasks, \sdrop provides a significant improvement over \textbf{Dense}, proving that it can indeed improve generalisation. When paired with the transformer architecture, \sdrop gives \emph{better} generalisation than \textbf{Dropout + Dense}. This is likely because hidden activations in transformers preserve locality information, hence, similar to how DropBlock \citep{drop-block} performs better with convolutional neural networks, \sdrop can more effectively regularise transformers. The end-to-end training time of \sdrop is also lower for the Fashion MNIST and Shakespeare tasks. In such cases, \sdrop is both \emph{better and faster than standard dropout}. I observe that the per-step convergence speed is similar between \sdrop and \textbf{Dropout + Dense}, thus most of the speed-up comes from the faster step time of \sdrop.  

At the same dropout rate $p$, \sdrop is a stronger regulariser than standard dropout. This is reflected in that the best $p$ found for \sdrop is lower than that for \textbf{Dropout + Dense} across all three tasks. Additionally, I observe that the training loss is higher for \sdrop than standard dropout for the same $p$. This matches our expectation as two consecutive entries of data are more likely to be correlated than two random entries, particularly when dealing with sequential/spatial inputs. As a result, dropping out consecutive blocks of data should destroy more information than randomly chosen items, leading to a stronger regularisation effect. From a computational perspective, this is a negative result as the preference for a smaller $p$ means there is less sparsity to exploit. 

\subsection{Real-world throughput} \label{sec:throughput}

\begin{figure}[t]
    \centering
    \subcaptionbox
    {ViT on Fashion MNIST (batch size 64).}
    [.41\textwidth]
    {\includegraphics[width=\linewidth]{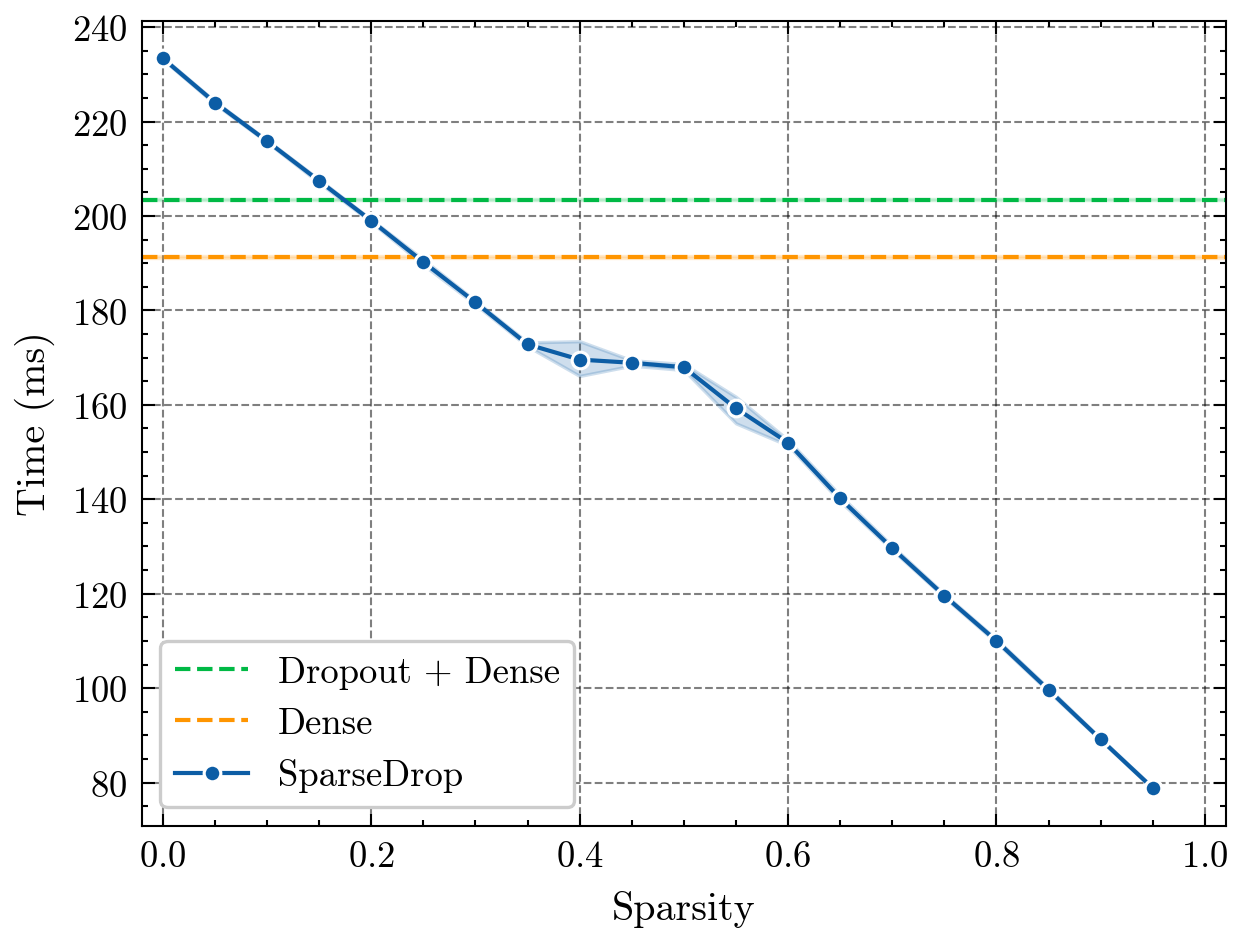}}%
    \qquad
    \subcaptionbox
    {GPT on Shakespeare (sequence length 128, batch size 32).}
    [.41\textwidth]
    {\includegraphics[width=\linewidth]{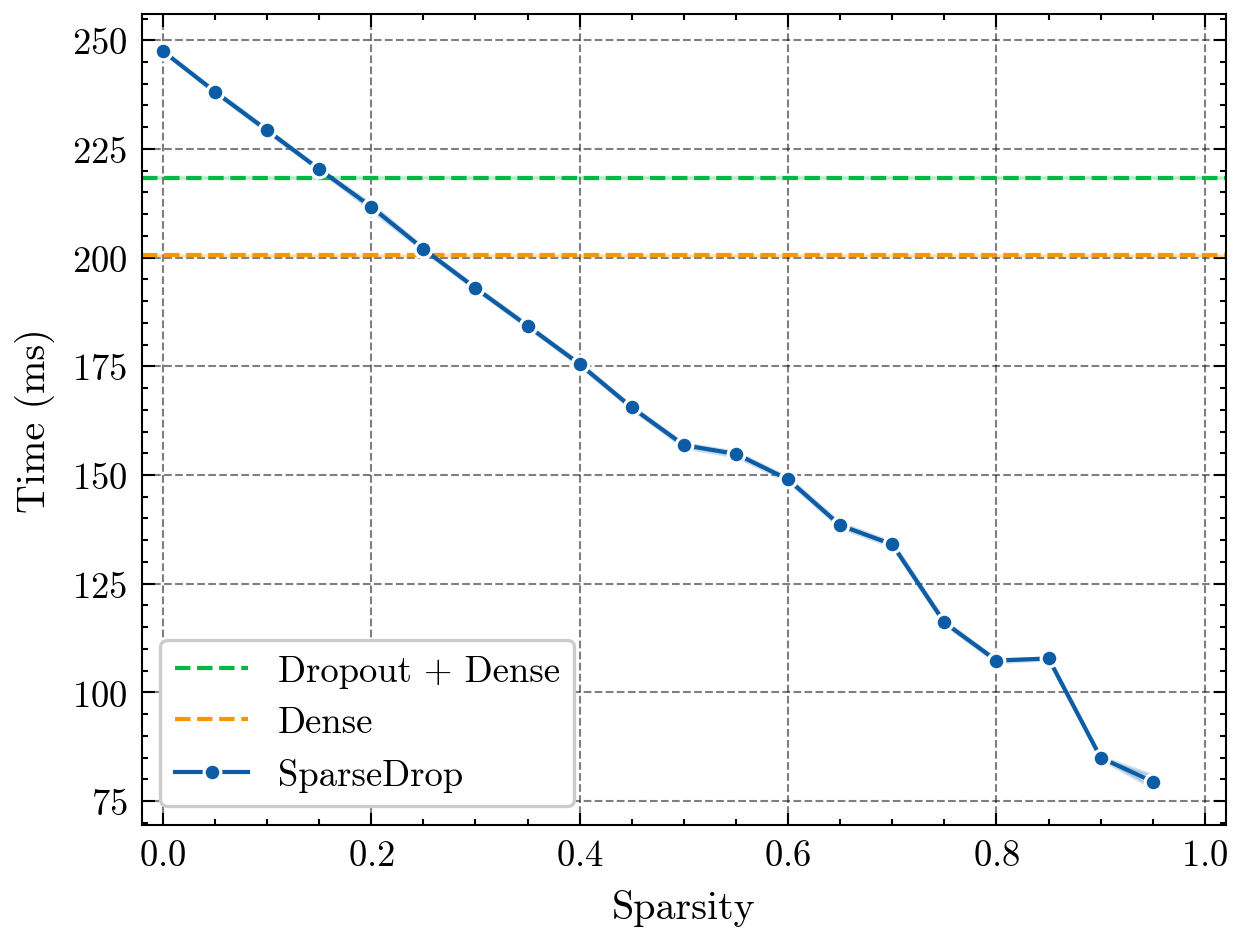}}%
    \caption{Total time (forward + backward) of models at various sparisty levels. Measured on RTX 2060 Max-Q (GPU clock locked at 1200MHz) with automatic mixed precision.}
    \label{fig:runtime}
\end{figure}

\cref{fig:runtime} shows the practical run times of full models during training. Similar to that observed in \cref{fig:benchmark-time}, the latency decreases linearly as the sparsity increases. With sparsity $\geq 20\%$, \sdrop already outperforms \textbf{Dropout + Dense}. We potentially achieve even higher speed-ups if we fully sparsify the model, such as using sparse operators for layers directly before dropout as well.

\section{Limitations and future directions}

\subsection{Performance tuning}

While \sdrop appears to be computationally efficient in an isolated benchmark (\cref{fig:benchmark}), I observe inconsistent timing behaviour when used in MLP. It is unclear what is the source of the issue though the expected linear-scaling trend is recovered with a larger model. Further engineering efforts are required to fix the issue.

There is still room to further optimise the GEMM performance. For example, all of the experiments use a block size of $128 \times 128$, even across different sparsity levels. I hypothesise that as the sparsity increases, smaller block sizes might be more optimal as it is effectively solving a smaller problem. \cref{sec:benchmark,sec:throughput} also reveal that there is still some performance gap between \sdrop and \textbf{Dense} when the sparsity is zero. I believe that with more engineering efforts (e.g., implementing software pipelining \citep{software-pipelining}) the gap can be completely closed.

\subsection{Hardware and software lock-in}

The implementation of \sdrop is heavily reliant on CUDA and is only developed for Turing architecture GPUs. This makes it difficult to deploy \sdrop on other accelerators (e.g., TPUs), or to achieve the same performance guarantees on a different generation of NVIDIA GPUs. Additionally, CUDA is a low-level programming language that takes considerable engineering effort to use. The implementation in this project is thus not easily extensible to different hardware. 

A higher-level kernel programming language like OpenAI Triton \citep{triton} might be preferred as it provides better cross-platform support. In fact, \sdrop is initially implemented in Triton. Unfortunately, I was unable to optimise the Triton implementation to match dense GEMM performance, possibly because Triton mainly targets Ampere or newer architectures. It is also not possible to inject custom C++ logic which is important for removing the mask generation bottleneck. Future work might want to revisit the Triton implementation as the Triton language matures. 

\subsection{Future directions}

While dropout is mostly used during training, it is occasionally also utilised for inference. A common application is Bayesian Dropout \citep{bayes-dropout}, where dropout serves as a source of stochasticity for Bayesian inference. \sdrop might be suitable for accelerating both the training and inference of such models. 

This project only focused on accelerating the `dropout followed by dense' pattern. While this has been proven to be successful in the experiments, it is more common to employ structured dropout for convolutional neural networks. There are potentially more gains to optimise for the `dropout followed by convolution' pattern, which in principle can be implemented in the same way as \sdrop with the implicit GEMM algorithm \citep{implicit-gemm}. 

\bibliography{references}

\appendix

\section{Experiment details} \label{sec:exp-details}

\subsection{MLP MNIST}

The full training configuration is:
\begin{minted}{json}
{
    "data": {
        "name": "mnist",
        "val_size": 4096,
        "train_size": 16384,
        "val_batch_size": 1024,
        "train_batch_size": 1024
    },
    "seed": 1,
    "model": {
        "dropout": {
            "p": 0.5,
            "variant": "blockwise[cuda]",
            "block_size": [
                128,
                128
            ]
        },
        "hidden_dim": 1024,
        "num_layers": 2,
        "output_dim": 10
    },
    "train": {
        "log_every": 10,
        "early_stop": {
            "mode": "max",
            "monitor": "Valiation accuracy",
            "patience": 5
        },
        "eval_every": 50,
        "max_epochs": 100
    },
    "wandb": {
        "mode": "online",
        "notes": null,
        "project": "flash-dropout-mlp"
    },
    "fabric": {
        "precision": "16-mixed",
        "accelerator": "auto"
    },
    "optimizer": {
        "lr": 0.001
    }
}
\end{minted}

\subsection{ViT Fashion MNIST}

The full training configuration is:
\begin{minted}{json}
{
    "data": {
        "name": "fashion_mnist",
        "val_size": 4096,
        "train_size": 16364,
        "val_batch_size": 64,
        "train_batch_size": 64
    },
    "seed": 0,
    "model": {
        "n_head": 8,
        "dropout": {
            "p": 0.5,
            "variant": "blockwise[cuda]",
            "block_size": [
                128,
                128
            ]
        },
        "n_embed": 1024,
        "n_layers": 2,
        "block_size": [
            8,
            8
        ],
        "patch_size": [
            2,
            2
        ]
    },
    "train": {
        "log_every": 50,
        "early_stop": {
            "mode": "max",
            "monitor": "Valiation accuracy",
            "patience": 5
        },
        "eval_every": 200,
        "max_epochs": 50
    },
    "wandb": {
        "mode": "online",
        "notes": null,
        "project": "flash-dropout-vit"
    },
    "fabric": {
        "precision": "16-mixed",
        "accelerator": "auto"
    },
    "optimizer": {
        "lr": 0.0001
    }
}
\end{minted}

\subsection{ViT CIFAR-10}

The full training configuration is:
\begin{minted}{json}
{
    "data": {
        "name": "cifar10",
        "val_size": 4096,
        "train_size": 16364,
        "val_batch_size": 64,
        "train_batch_size": 64
    },
    "seed": 0,
    "model": {
        "n_head": 8,
        "dropout": {
            "p": 0.4,
            "variant": "blockwise[cuda]",
            "block_size": [
                128,
                128
            ]
        },
        "n_embed": 1024,
        "n_layers": 2,
        "block_size": [
            8,
            8
        ],
        "patch_size": [
            2,
            2
        ]
    },
    "train": {
        "log_every": 50,
        "early_stop": {
            "mode": "max",
            "monitor": "Valiation accuracy",
            "patience": 10
        },
        "eval_every": 200,
        "max_epochs": 50
    },
    "wandb": {
        "mode": "online",
        "notes": null,
        "project": "flash-dropout-vit"
    },
    "fabric": {
        "precision": "16-mixed",
        "accelerator": "auto"
    },
    "optimizer": {
        "lr": 0.0001
    }
}
\end{minted}

\subsection{LLM Shakespeare}

The full training configuration is:
\begin{minted}{json}
{
    "data": {
        "name": "shakespeare",
        "length": 128,
        "val_size": 1024,
        "cache_dir": "data/shakespeare",
        "train_size": 4096,
        "val_batch_size": 64,
        "train_batch_size": 32
    },
    "seed": 0,
    "model": {
        "n_head": 8,
        "dropout": {
            "p": 0.4,
            "variant": "blockwise[cuda]",
            "block_size": [
                128,
                128
            ]
        },
        "n_embed": 1024,
        "n_layer": 4,
        "context_length": 128
    },
    "train": {
        "log_every": 50,
        "early_stop": {
            "mode": "min",
            "monitor": "Valiation loss",
            "patience": 5
        },
        "eval_every": 200,
        "max_epochs": 100
    },
    "wandb": {
        "mode": "online",
        "notes": null,
        "project": "flash-dropout-llm"
    },
    "fabric": {
        "precision": "16-mixed",
        "accelerator": "auto"
    },
    "optimizer": {
        "lr": 0.0003,
        "weight_decay": 0.1
    }
}
\end{minted}

\end{document}